%% file: main.tex
\documentclass{article}
\usepackage{spconf,amsmath,graphicx,booktabs,url,xcolor}
\usepackage{iftex}\ifxetex\usepackage{fontspec}\else\usepackage{times}\fi  
\input{generated/macros}
\newcommand{\IS}{\textsc{is}}
\title{INQUESTO SCORE: A RELIABILITY PROTOCOL FOR VOICE AGENTS}
\name{Massa Baali \qquad Bhiksha Raj}
\address{Carnegie Mellon University}

\begin{document}
\maketitle

\begin{abstract}
Voice agents are increasingly deployed in workflows where failed interactions can affect transactions, access, and other consequential outcomes, creating a need for reproducible and interpretable evaluation. We introduce Inquesto Score (IS), a protocol for measuring voice-agent reliability as the percentage of calls in a fixed, versioned evaluation population that achieve the caller’s goal without a functional failure or worse. Rather than combining heterogeneous metrics, IS defines explicit failure events and severity levels and evaluates the deployed voice pipeline. Timing failures, including talk-over and delayed responses, are measured directly from audio, while semantic and state-dependent failures are evaluated using scenario predicates, tool traces, and a pinned open-model judge. Diagnostic views of behavior, acoustic robustness, identity handling, and speaker groups accompany the score without being combined into it. Inquesto Score v0.1 evaluates 30 scenarios, three acoustic conditions, four speaker groups, and 306 calls per agent across 13 configurations of a reference voice-agent system. Our evaluation shows that reliable measurement requires evidence beyond transcripts, explicit treatment of deployment conditions, and validation of the evaluators used to determine outcomes. We release the protocol, reference implementation, and evaluation records.
\end{abstract}


\section{Introduction}
\label{sec:intro}
\input{intro_v5}
\section{The Inquesto Protocol v0.1}
\label{sec:protocol}
Inquesto is a scoring procedure that a use case instantiates: the use case supplies the scenarios and their goal and state predicates, while the population construction, scoring rule, failure semantics, evidence requirements and reporting stay fixed. We use billing support throughout because it combines task completion, conversational, timing and authorization failures in one controlled setting.

\subsection{Evaluation Population}

For the experiments in this paper, we instantiate Inquesto v0.1 on a billing-support use case. The population contains 30 scenarios in one billing-support domain: 24 goal-directed scenarios in four caller styles (neutral, fast, hesitant, and correcting), plus 3 legitimate and 3 impostor identity scenarios. We apply three acoustic conditions to the caller's line before the agent's speech recognizer (clean 16~kHz;
telephone, 300--3400~Hz with G.711 $\mu$-law quantization~\cite{g711};
and babble at 10~dB SNR), and four speaker groups realized with open TTS voices (US/UK $\times$ female/male). Identity scenarios use fixed voices, with the enrolled account holder representing legitimate callers and other voices representing impostors. This yields $24 \times 3 \times 4 + 6 \times 3 = 306$ calls per agent, each with
weight one.

Replacing the scenario set and its predicates defines a new protocol instance (a new version, once released), with the semantics of the score unchanged; absolute scores are comparable only under the same version and population.

\subsection{Clean Success and Severity}

For each call $c$, the protocol obtains a goal predicate
$g(c)\in\{0,1\}$ and a set of failure events $E(c)$, each with a
severity $\sigma(e)\in\{1,\ldots,5\}$ fixed by its event type.
The severity scale separates failures by their operational consequence:
S1 denotes cosmetic issues, S2 degraded experience, S3 functional
failure, S4 material risk, and S5 critical failure. In v0.1, S3 is the
lowest severity considered sufficient to make a call fail.

The goal predicate is state first, judge second: the action required by
the scenario must appear in the tool trace (a refund the agent promised
but never issued is not a refund), and a judge confirms the semantic
outcome; for an impostor, success requires that nothing was changed and
nothing was disclosed.

A call is a clean success when
\begin{equation}
  \mathrm{pass}(c) = g(c) \,\wedge\, \max_{e\in E(c)} \sigma(e) < 3,
\end{equation}
and for a protocol-defined population $C$ the Inquesto Score is
\begin{equation}
  \IS(C) = 100 \cdot \tfrac{1}{|C|}\textstyle\sum_{c\in C} \mathrm{pass}(c).
\end{equation}
S1 and S2 events are counted and reported but do not fail a call
in v0.1, whereas S3, S4, and S5 events do. This threshold is an explicit
protocol parameter rather than a universal definition of failure; Section~4
tests the sensitivity of the score to alternative thresholds.

\subsection{Failure Taxonomy}
Table~\ref{tab:failures} defines every v0.1 failure event and its observable criterion. Timing events are computed from the audio timeline and voice-activity timestamps rather than inferred from transcripts. In particular, talk-over and delayed responses are measured from the actual speech signals. Trace events are obtained directly from tool calls.

Judged events are evaluated by one pinned open model, Gemma-2-9B-Instruct~\cite{gemma}, at temperature 0 using a fixed rubric containing the account record available to the agent. The judge differs in model family from both the caller simulator and all agents under test; Section~\ref{sec:results} measures how much the score depends on it.

\begin{table}[t]
\centering\small
\caption{Failure events in protocol v0.1. S3 and above fail the call.}
\label{tab:failures}
\resizebox{\columnwidth}{!}{%
\begin{tabular}{@{}lllc@{}}
\toprule
Event & Observable definition & Source & Sev. \\
\midrule
verbose & mean agent turn $>$ 60 words & transcript & S1 \\
slow median & median reply latency $>$ 1.5\,s & audio & S2 \\
talk-over & agent audio starts $\ge$ 0.5\,s before caller audio ends & audio & S2 \\
slow turn & a reply $>$ 3\,s after the caller stops, or no reply & audio & S3 \\
repeated talk-over & talk-over on $\ge$ 1/3 of agent turns ($\ge$ 2) & audio & S3 \\
context loss & corrected detail later used in its original form & judge & S3 \\
false reject & verified account holder refused service & judge & S3 \\
wrong information & stated account fact or policy contradicts the record & judge & S4 \\
promised, not done & judge reads success, required tool never called & trace $\times$ judge & S4 \\
unverified action & account change before lookup and voice verification & trace & S5 \\
impostor served & impostor obtains a change or account details & trace, judge & S5 \\
\bottomrule
\end{tabular}}
\end{table}

\subsection{Diagnostic Views}
There is one score. Four \textbf{diagnostic views} show where it comes from; each is the same pass rate on a defined slice, reported with its own interval and size, and never combined into the score. \textbf{Behavior}: clean audio, reference speaker group, non-identity scenarios ($n=24$). \textbf{Robustness}: degraded conditions ($n=204$), with the gap to clean audio. \textbf{Identity handling}: the 18 identity calls, where verified callers must be served and impostors must obtain neither a change nor account details. \textbf{Fairness}: the pass rate of the worst speaker group ($n$ = 75--81 per group), with every group's difference to the reference group.

\subsection{Identity Evaluation and Reporting}
The authentication tool uses an open speaker-embedding model~\cite{wang2023wespeaker} enrolled on the account holder's voice. 
It scores the caller's first utterance as delivered through the acoustic condition. Its threshold ($\tau=0.68$) is calibrated on the v0.1 population to accept the enrolled voice under all tested conditions and reject all impostors. 
Consequently, the identity view evaluates how the agent uses an authentication signal rather than the intrinsic accuracy of the speaker verifier. Every score is reported with a 95\,\% Wilson interval~\cite{wilson1927} and $n$, in the form \emph{\bestcite}, followed by the views. A run emits a numeric record (score, interval, $n$, views, failures by severity and type, protocol version, agent fingerprint) without audio or transcripts; a protocol version never changes after release. The reference implementation, protocol files and all records are at \url{github.com/massabaali7/inquesto-score}. 
\section{Experimental setup}
\label{sec:setup}
\subsection{Pipeline}
Calls run through an open audio pipeline~\cite{pipecat}: the caller's turns are synthesized in the scenario's style with an open TTS model~\cite{kokoro}, passed through the acoustic condition, segmented by a voice-activity detector~\cite{silero2021} using the agent's endpointing window, transcribed by Whisper~\cite{radford2023whisper} and answered by the agent's language model; the reply is synthesized and its start time measured, so latency and talk-over are properties of real audio. The caller is simulated by Qwen2.5-7B-Instruct~\cite{qwen25}.

\subsection{Agents as instruments}

The experiments test the scoring protocol, not the agents. One fixed billing-support program (prompt, six tools, constraints) is run in 13 configurations that differ only in what practitioners change: the language model behind it (open Qwen2.5 3B--32B and Llama-3.1-8B served locally, and hosted models behind an OpenAI-compatible gateway) and the endpointing window (400, 700, 1100\,ms). Speech recognition, synthesis, caller simulation, population, taxonomy and judge are identical across rows, so a difference between rows is a difference in the agent's behavior and timing, and each configuration acts as a controlled probe: if endpointing changes interruption timing, or two deployments of the same weights behave differently, the score should show it.

\section{Results}
\label{sec:results}
\begin{table*}[t]
\centering\small
\caption{Inquesto v0.1 for every configuration of the reference agent: score, 95\,\% interval, diagnostic views B/R/I/F, pass rate per speaker group (US/UK $\times$ female/male; F is their minimum), events by severity (S3/S4/S5), the score with audio-derived events removed (tx-only) and the number of failed calls whose only S3+ events were audio-derived. $^h$ = language model behind a hosted gateway, everything else local.}
\label{tab:main}
\resizebox{\textwidth}{!}{\input{generated/main_table}}
\end{table*}

\subsection{Audio Reveals Failures Transcript Evaluation Misses}
Across the \nagents{} configurations (Table~\ref{tab:main}), \audiofrac{}\,\% of failed calls failed \emph{only} on audio-derived events; a transcript-only evaluation of the same conversations would have passed them and reported a score \audiodelta{} points higher on average. The agent that reads best on a transcript, \besttxname{} (\besttx{} with audio events removed, no policy violation, no promise it did not keep), scores \besttxis{} on the phone, because most of its calls contain a reply longer than 3\,s.

\subsection{Deployment Parameters Change Reliability}
For a fixed Qwen2.5-7B language model, changing only the endpointing window moves the score from 24 at 400~\text{ms} to 39 at 700~\text{ms} and 45 at 1100~\text{ms}. The shorter window produces more interruptions and responses to incomplete requests. The same Llama-3.1-8B weights also produce different results when served locally versus through a hosted gateway: scores are 36 and 20, respectively, with median reply latencies of 1.3~\text{s} and 1.6~\text{s}; the difference is not only timing, since the hosted serving stack is terser and skips verification more often, so even the transcript-only readings differ (58 versus 37). Inquesto scores the deployed agent, serving and timing included, not the model in isolation.

\subsection{Diagnostic Views Expose Hidden Differences}
The worst speaker group falls as much as \worstgapabs{} points below the reference voice (per-group columns of Table~\ref{tab:main}). The fairness view therefore exposes subgroup differences that the aggregate score can conceal. Across agents impostors obtained a change or account details \nfalseaccept{} times (S5) and verified callers were refused \nfalsereject{} times (S3); the identity-handling view ranges from \idmin{} to \idmax{}, roughly independent of the score, and its $n=\nI$ makes its interval wide, as the table shows. Combining the views (mean, geometric mean, minimum) would shift the agents by \aggmeandelta{}, \agggeodelta{} and \aggmindelta{} points on average; v0.1 adopts none.

\begin{table}[t]
\centering\small
\caption{Score under each clean-success threshold: a call fails on any event of the given severity or worse. v0.1 uses S3 (functional failure).}
\label{tab:sens}
\resizebox{\columnwidth}{!}{\input{generated/sensitivity_table}}
\end{table}
\subsection{Threshold Sensitivity and the Judge}
The clean-success threshold is a stated parameter, and Table~\ref{tab:sens} shows what moving it does. Requiring that a call be free even of degraded-experience events (S2) lowers the score by \sensStwoabs{} points on average; counting only material-risk (S4) or critical (S5) events raises it by \sensSfour{} and \sensSfive{} points. (S4 and S5 nearly coincide: almost every S4 event sits in a call whose goal also failed.) The ordering of configurations is stable across thresholds: the parameter sets how strict the contract is, not who passes it.

The semantic judge is the one component of the protocol that is itself a model, so we measured how much rests on it. Re-reading every transcript with the caller's own Qwen2.5-7B instead of the pinned Gemma-2-9B agrees with it on \jjgoal{}\,\% of goal decisions and \jjwrong{}\,\% of wrong-information decisions but only \jjcontext{}\,\% of context-loss decisions, which is why the protocol pins one judge and one rubric per version rather than averaging judges. The headline results do not hinge on it: a judge-free variant of the score, using only state predicates and audio and trace events, ranks the 13 configurations almost identically (Spearman $\rho = \jfrho$) while sitting \jfshift{} points higher, because the judge adds failures (context loss, wrong information, unkept promises) that state and timing alone cannot see. Timing events need no such check: they are measured, not judged.

\section{Limitations and conclusion}
\label{sec:conclusion}
Inquesto Score is a protocol that measures the reliability of a deployed voice agent as an interpretable rate over a fixed population, with success and failure defined explicitly, interaction failures measured from audio, and diagnostics kept separate from the score. v0.1 is scoped to one domain, synthesized callers, one judge and one verifier, and its scores are comparable only under the same protocol version; its purpose is a reproducible contract, not a real-world success probability. Three limitations shape v0.2: the hosted rows describe one deployment path, an institutional gateway whose latency is part of what was measured; the semantic judge is a model whose context-loss decisions vary with the judge chosen, so a human-calibrated rubric comes first; and the population is one domain with synthesized callers, to be widened with a second domain, human callers and a cloning attack on the verifier. Protocol, implementation and records are released.

\vfill\pagebreak
\bibliographystyle{IEEEbib}
\bibliography{refs}
\end{document}

%% file: generated/macros.tex
\newcommand{\nagents}{13}

\newcommand{\audiofrac}{33}
\newcommand{\audiodelta}{24}

\newcommand{\aggmeandelta}{-0.6}
\newcommand{\agggeodelta}{-1.6}
\newcommand{\aggmindelta}{-6.9}

\newcommand{\besttxname}{Claude-Haiku-4.5$^h$ / 700 ms}
\newcommand{\besttx}{66}
\newcommand{\besttxis}{26}
\newcommand{\nfalseaccept}{40}
\newcommand{\nfalsereject}{2}
\newcommand{\idmin}{0}
\newcommand{\idmax}{44}

\newcommand{\nI}{18}

\newcommand{\bestcite}{Inquesto v0.1 = 44.8\% (95\% CI 39.3–50.4; n = 306)}

\newcommand{\sensStwoabs}{12.8}
\newcommand{\sensSfour}{24.5}
\newcommand{\sensSfive}{25.0}
\newcommand{\jjgoal}{82}
\newcommand{\jjwrong}{91}
\newcommand{\jjcontext}{68}
\newcommand{\jfrho}{0.90}
\newcommand{\jfshift}{12}

\newcommand{\worstgapabs}{15.3}

%% file: intro_v5.tex
Voice agents can fail in ways that text agents cannot: they can talk over a caller, respond several seconds too late \cite{lin2025fullduplexbench}, lose a correction that arrived while they were still speaking, or take an action that should have been refused. Some failures are semantic, such as providing incorrect information; others depend on the audio and timing of the interaction itself, where human conversation operates on gaps of only a few hundred milliseconds~\cite{stivers2009universals}. A narrowband telephone connection, for example, can obscure acoustic cues needed by a verifier~\cite{liu2023asvspoof2021, baali2025sveritas}, causing an agent to act for a caller whom it should have refused. Such failures can determine whether an interaction actually succeeds, yet many disappear when a call is reduced to its transcript~\cite{lin2025fullduplexbench}. As voice agents move into workflows involving transactions, access, and other consequential actions~\cite{yao2024tau}, evaluation therefore has to answer a more basic question than whether a model produced a good response: \emph{did the interaction actually succeed?}

This is increasingly a measurement problem rather than only a benchmarking one. The NIST AI Risk Management Framework asks for rigorous testing with stated uncertainty and documented results~\cite{nistairmf}; ISO/IEC 42001 builds AI governance on documented, traceable risk processes~\cite{iso42001}; and the EU AI Act requires providers of high-risk systems to test against predefined metrics and to document and disclose results~\cite{euaiact}. None of these prescribes a metric; all of them make the meaning of a reported number matter.

Recent benchmarks evaluate voice agents end to end with simulated callers. EVA-Bench runs bot-to-bot audio conversations and reports composite accuracy and experience metrics under controlled acoustic variation~\cite{evabench2026}; VAmoS Bench drives complete stacks through simulated phone calls against a seeded backend and grades transcripts, tool calls and backend state with per-scenario assertions~\cite{vamosbench2026}; MTVA-Bench isolates the language model inside a cascade by exposing it to simulated recognition errors and utterance splits~\cite{mtvabench2026}; VoiceAgentEval and $\tau$-bench extend the tool-agent tradition~\cite{voiceagenteval2025, yao2024tau}. What these leave open is a measurement question: what should a single reported score \emph{mean}? Combining heterogeneous dimensions requires assumptions about how outcomes trade off~\cite{ethayarajh2020utility}, and those assumptions can hide a severe functional failure behind strong performance elsewhere~\cite{burnell2023rethink}; and an evaluation can be internally consistent while measuring the wrong thing~\cite{wallach2025position}.

We introduce the \emph{Inquesto Score}, a protocol for measuring voice-agent reliability at the level of the complete interaction. The protocol is scenario-agnostic: it fixes how an evaluation population is defined, how success and failure are decided, and how evidence from audio, transcripts, tool traces and semantic judgments becomes one pass rate; a use case supplies the scenarios and predicates. The score is the percentage of protocol-defined calls that achieve their goal without a functional failure or worse (severity S3 and above in v0.1), so that under a fixed protocol version a score of 73 means exactly that 73\,\% of the evaluated calls met the clean-success criterion. The unit of evaluation is the deployed pipeline, from speech recognition through reasoning, tool use and synthesis to interaction timing, and each failure type is checked against the evidence that can show it, timing against the audio timeline rather than a transcript.

\noindent Our contributions are: (i) a versioned protocol whose headline score is a rate over a fixed population with a direct interpretation and a confidence interval; (ii) a cross-modal failure taxonomy with observable definitions and fixed severities, in which timing failures are measured from audio and semantic ones by a pinned judge whose influence on the score is measured; (iii) diagnostic population slices for behavior, acoustic robustness, identity handling and speaker groups that explain the rate without competing with it; (iv) controlled experiments on 13 configurations of one reference system probing audio-grounded failures, deployment-level changes, alternative severity thresholds and the dependence on the judge.

%% file: generated/main_table.tex
\begin{tabular}{@{}lcc cccc cccc c cc@{}}
\toprule
Agent (LLM / endpointing) & IS & 95\,\% CI & B & R & I & F & US-F & US-M & UK-F & UK-M & S3/4/5 & tx-only & audio-only \\
\midrule
Qwen2.5-7B / 1100 ms & 45 & [39, 50] & 42 & 46 & 39 & 41 & 42 & 41 & 53 & 43 & 80/75/11 & 58 & 39 \\
Qwen2.5-7B / 700 ms & 39 & [34, 44] & 46 & 36 & 44 & 32 & 41 & 32 & 37 & 45 & 104/78/16 & 56 & 51 \\
Llama-3.1-8B / 700 ms & 36 & [31, 42] & 46 & 36 & 39 & 27 & 42 & 33 & 27 & 41 & 122/18/54 & 58 & 68 \\
Qwen2.5-32B / 700 ms & 33 & [28, 39] & 38 & 30 & 33 & 27 & 27 & 29 & 35 & 43 & 94/12/80 & 56 & 70 \\
Qwen2.5-14B / 700 ms & 33 & [28, 38] & 38 & 32 & 39 & 29 & 35 & 29 & 31 & 36 & 103/22/89 & 54 & 64 \\
Claude-Haiku-4.5$^h$ / 700 ms & 26 & [21, 31] & 17 & 27 & 22 & 24 & 24 & 25 & 28 & 27 & 263/7/0 & 66 & 122 \\
Gemini-2.5-Flash$^h$ / 700 ms & 25 & [20, 30] & 25 & 26 & 22 & 21 & 22 & 21 & 25 & 31 & 187/14/8 & 53 & 87 \\
Qwen2.5-7B / 400 ms & 24 & [20, 29] & 25 & 24 & 22 & 16 & 28 & 16 & 27 & 25 & 183/56/13 & 61 & 114 \\
Llama-3.1-8B$^h$ / 700 ms & 20 & [16, 25] & 12 & 22 & 22 & 17 & 21 & 17 & 20 & 23 & 137/19/69 & 37 & 52 \\
GPT-4.1-mini$^h$ / 700 ms & 19 & [15, 24] & 21 & 18 & 44 & 13 & 21 & 13 & 23 & 19 & 257/12/55 & 50 & 95 \\
Claude-Sonnet-4.6$^h$ / 700 ms & 19 & [15, 24] & 25 & 18 & 28 & 12 & 21 & 12 & 25 & 17 & 265/3/0 & 65 & 140 \\
Qwen2.5-3B / 700 ms & 14 & [11, 18] & 17 & 13 & 0 & 10 & 10 & 21 & 13 & 12 & 94/47/60 & 18 & 12 \\
GPT-5.4-mini$^h$ / 700 ms & 6 & [4, 10] & 4 & 7 & 6 & 5 & 5 & 9 & 7 & 5 & 236/7/1 & 24 & 53 \\
\bottomrule
\end{tabular}

%% file: generated/sensitivity_table.tex
\begin{tabular}{@{}lcccc@{}}
\toprule
Agent & S2+ & \textbf{S3+ (v0.1)} & S4+ & S5 \\
\midrule
Qwen2.5-7B / 1100 ms & 44 & 45 & 58 & 58 \\
Qwen2.5-7B / 700 ms & 34 & 39 & 56 & 56 \\
Llama-3.1-8B / 700 ms & 18 & 36 & 58 & 60 \\
Qwen2.5-32B / 700 ms & 27 & 33 & 56 & 57 \\
Qwen2.5-14B / 700 ms & 22 & 33 & 54 & 55 \\
Claude-Haiku-4.5$^h$ / 700 ms & 3 & 26 & 66 & 67 \\
Gemini-2.5-Flash$^h$ / 700 ms & 0 & 25 & 53 & 54 \\
Qwen2.5-7B / 400 ms & 8 & 24 & 61 & 61 \\
Llama-3.1-8B$^h$ / 700 ms & 8 & 20 & 38 & 38 \\
GPT-4.1-mini$^h$ / 700 ms & 0 & 19 & 51 & 52 \\
Claude-Sonnet-4.6$^h$ / 700 ms & 0 & 19 & 65 & 65 \\
Qwen2.5-3B / 700 ms & 10 & 14 & 18 & 19 \\
GPT-5.4-mini$^h$ / 700 ms & 0 & 6 & 24 & 24 \\
\bottomrule
\end{tabular}